\documentclass[10pt,twocolumn,letterpaper]{article}

\usepackage{iccv}
\usepackage{times}
\usepackage{epsfig}
\usepackage{graphicx}
\usepackage{amsmath}
\usepackage{amssymb}

\usepackage[breaklinks=true,bookmarks=false]{hyperref}

\iccvfinalcopy 

\begin{document}

\title{Dense Scale Network for Crowd Counting}

\author{Feng Dai$^{1}$, Hao Liu$^{1,2}$, Yike Ma$^{1}$, Juan Cao$^{1}$, Qiang Zhao$^{1*}$, Yongdong Zhang$^{3}$\\
$^{1}$Institute of Computing Technology, Chinese Academy of Sciences, Beijing, China\\
$^{2}$University of Chinese Academy of Sciences, Beijing, China\\
$^{3}$University of Science and Technology of China, Hefei, China\\
{\tt\small \{fdai,liuhao2018,ykma,caojuan,zhaoqiang\}@ict.ac.cn}  
{\tt\small zhyd73@ustc.edu.cn}}


\maketitle

\begin{abstract}
Crowd counting has been widely studied by computer vision community in recent years. Due to the large scale variation, it remains to be a challenging task. Previous methods adopt either multi-column CNN or single-column CNN with multiple branches to deal with this problem. However, restricted by the number of columns or branches, these methods can only capture a few different scales and have limited capability. In this paper, we propose a simple but effective network called DSNet for crowd counting, which can be easily trained in an end-to-end fashion. The key component of our network is the dense dilated convolution block, in which each dilation layer is densely connected with the others to preserve information from continuously varied scales. The dilation rates in dilation layers are carefully selected to prevent the block from gridding artifacts. To further enlarge the range of scales covered by the network, we cascade three blocks and link them with dense residual connections. We also introduce a novel multi-scale density level consistency loss for performance improvement. To evaluate our method, we compare it with state-of-the-art algorithms on four crowd counting datasets (ShanghaiTech, UCF-QNRF, UCF\_CC\_50 and UCSD). Experimental results demonstrate that DSNet can achieve the best performance and make significant improvements on all the four datasets (30\% on the UCF-QNRF and UCF\_CC\_50, and 20\% on the others).
\end{abstract}

\section{Introduction}

With the rapid growth of population, crowd counting has gained considerable attention in recent years, because of its broad applications such as video surveillance, traffic control and sport events. Earlier works estimate crowd counts via the detection of body or head \cite{rodriguez2011density,wang2011automatic,wu2005detection}, while some other methods learn a mapping from local or global hand-crafted feature to the actual count \cite{ryan2009crowd,chan2009bayesian,chan2008privacy}. More recently, the problem of crowd counting is formalized into the regression of crowd density map, whose values are summed to give the count of crowd within that image. This approach can handle serious occlusions in dense crowd images. With the success of deep learning technology, convolutional neural networks (CNNs) are utilized to generate accurate crowd density map and can get better performance than traditional methods.

\begin{figure}
\begin{center}
\begin{minipage}{4cm}
\includegraphics[width=4cm, height=2.7cm]{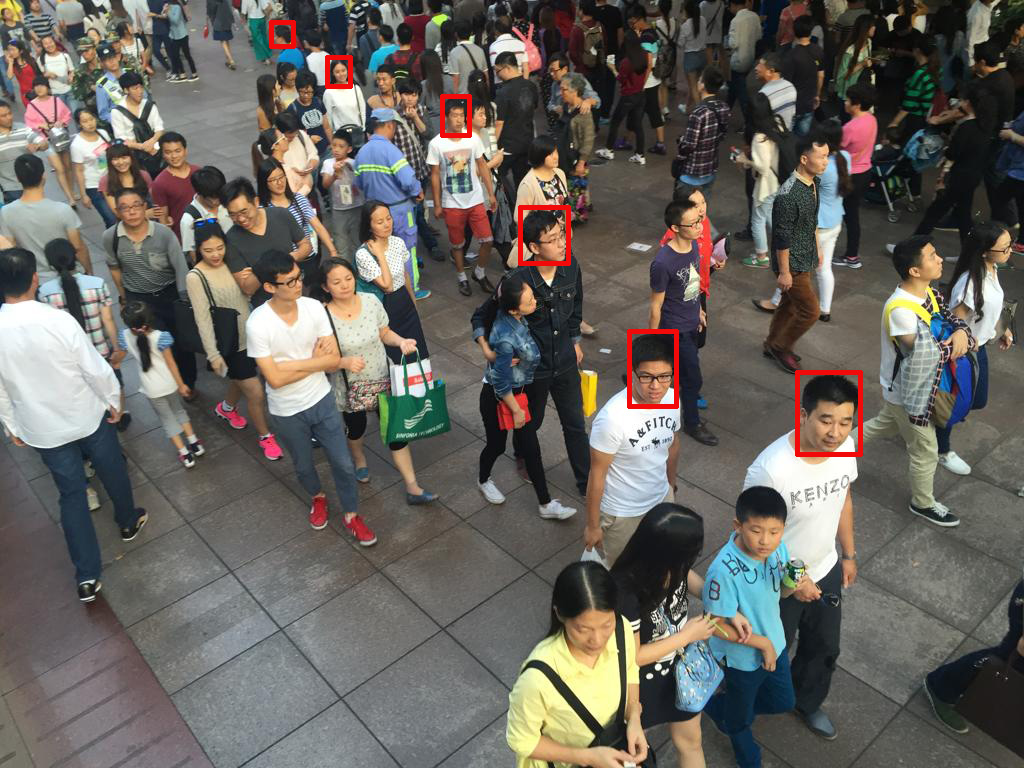}
\end{minipage}
\begin{minipage}{4cm}
\includegraphics[width=4cm, height=2.7cm]{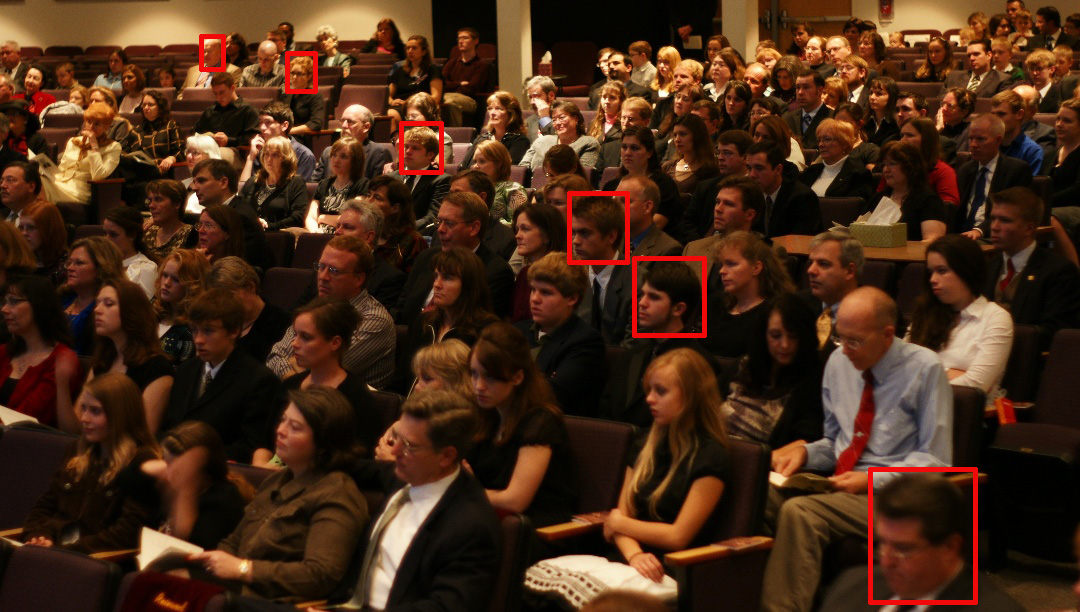}
\end{minipage}

\begin{minipage}{4cm}
\includegraphics[width=4cm, height=2.7cm]{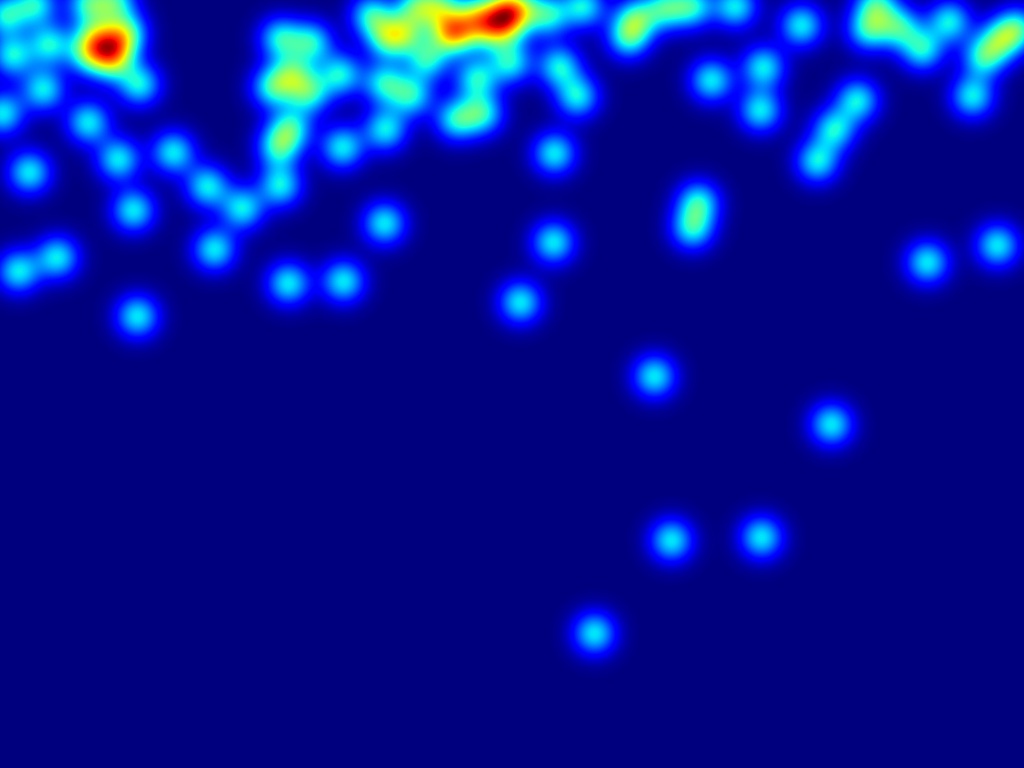}
\end{minipage}
\begin{minipage}{4cm}
\includegraphics[width=4cm, height=2.7cm]{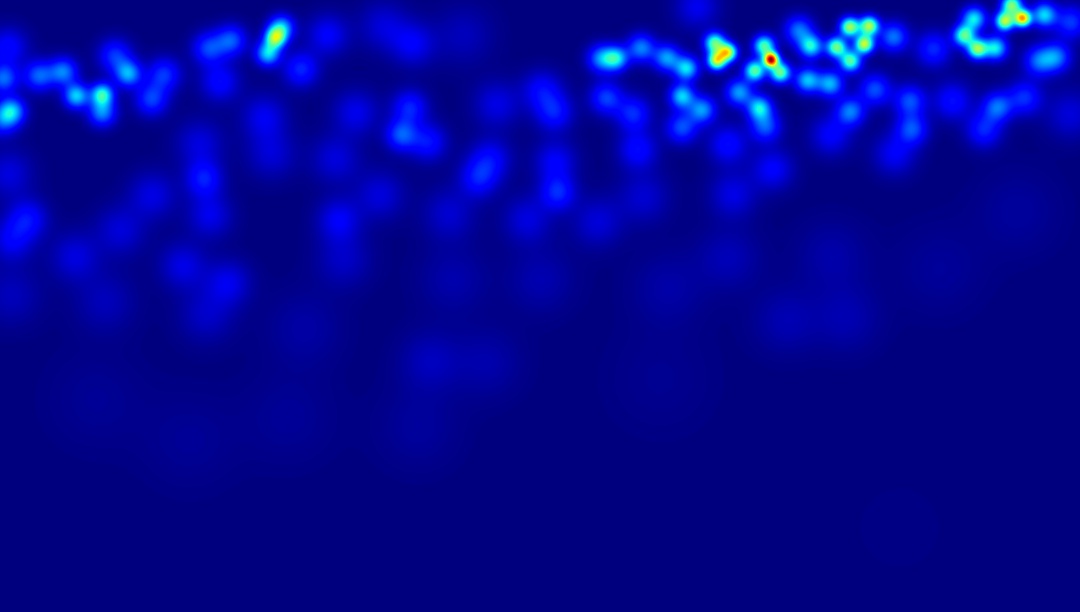}
\end{minipage}

\end{center}
\caption{Large scale variations exist in crowd counting datasets. Left: Input image and corresponding ground truth density map from ShanghaiTech dataset \cite{zhang2016single}. Right: Input image and corresponding ground truth density map from UCF-QNRF dataset \cite{idrees2018composition}.}
\label{fig1}
\end{figure}

However, crowd counting still remains to be an extremely challenging task due to large scale variation, heavy occlusion, background noise and perspective distortion. Among them, scale variation is the main issue and has attracted the most attention of recent CNN-based methods \cite{boominathan2016crowdnet,zhang2016single,onoro2016towards,sam2017switching,sindagi2017cnn,zhang2018crowd,cao2018scale}. A number of multi-column or multi-branch networks have been proposed to handle scale variation for better accuracy. These architectures contain several columns of CNN or several branches from different stages of backbone network. The columns or branches have different receptive fields to capture variation in people sizes. Although these methods show good improvements, the scale diversity they captured is restricted by the number of columns or branches. 

The key challenges of scale variation lie in two aspects. First, as shown in Figure \ref{fig1} left, the people in crowd image often have very different sizes, ranging from several pixels to tens of pixels. This requires the network to be able to capture a \textit{large rang} of scales. Second, as shown in Figure \ref{fig1} right, the scale usually varies continuously across the image, especially for high density images. This requires the network to be able to sample the captured scale range \textit{densely}. However, \textit{non} of existing methods can deal with these two challenges simultaneously. 


In this paper, we propose a novel dense scale single-column neural network called DSNet for crowd counting. DSNet is built upon the blocks consisting of densely connected dilated convolutional layers. Thus it can output features having different receptive fields and capture crowds at different scales. The block has similar structure with DenseASPP \cite{yang2018denseaspp} used for semantic segmentation, but with different combination of dilation rates. We carefully select these rates for layers within the blocks, such that each block can \textit{more densely} sample the continuously varied scale than DenseASPP. At the same time, the selected combination of dilation rates can use \textit{all} pixels under the receptive field for feature computation, and let us step from gridding artifacts. To further increase the scale diversity captured by DSNet, we stack three dense dilated convolution blocks and link them by utilizing the residual connection \cite{he2016deep} densely. The final network can sample a very large scale range in a much denser manner, thus has the ability to deal with the large scale variation problem in crowd counting.

Another issue of most previous methods is that they use traditional Euclidean loss to train their networks, which is based on the pixel independence hypothesis. This loss ignores the global and local coherence in density maps and would give poor performance in crowd counting. To overcome this issue, we incorporate a novel multi-scale density level consistency loss, which is used to ensure the global and local density level consistency between the estimated and ground truth crowd density map. The new proposed loss function can be quickly calculated and further improve the performance of DSNet.

In summary, the main contributions of our paper are:
\begin{itemize}
	\item We propose the dense dilated convolution block (DDCB) with carefully selected dilation rates, such that DDCB can more densely sample the continuously varied scale and step from gridding artifacts. We stack and link three DDCBs via dense residual connection to enlarge the range of scales. The final network, i.e. DSNet, can be easily trained end-to-end and can deal with both congested and sparse crowd images.
	
	\item Besides of Euclidean loss that only cares about pixel-wise error, we additionally introduce a multi-scale density level consistency loss for performance improvement. This loss enforces the global and local consistency between the estimated and ground truth density maps according to density levels.
	
	\item We carry out extensive experiments on four public challenging crowd counting datasets. Our approach can achieve the best performance compared with existing state-of-the-art methods. Especially, remarkable improvements (around 30\%) are presented on UCF-QNRF and UCF\_CC\_50 datasets. We also observe significant improvements (around 20\%) on ShanghaiTech and UCSD datasets.
\end{itemize}
\section{Related Work}

The existing crowd counting methods can be classified into two categories: traditional counting methods and CNN-based methods. By combining the deep learning, the CNN-based counting methods demonstrate the powerful ability for crowd counting and outperform the traditional methods.

\begin{figure*}[t]
\begin{center}
   \includegraphics[width=1\linewidth]{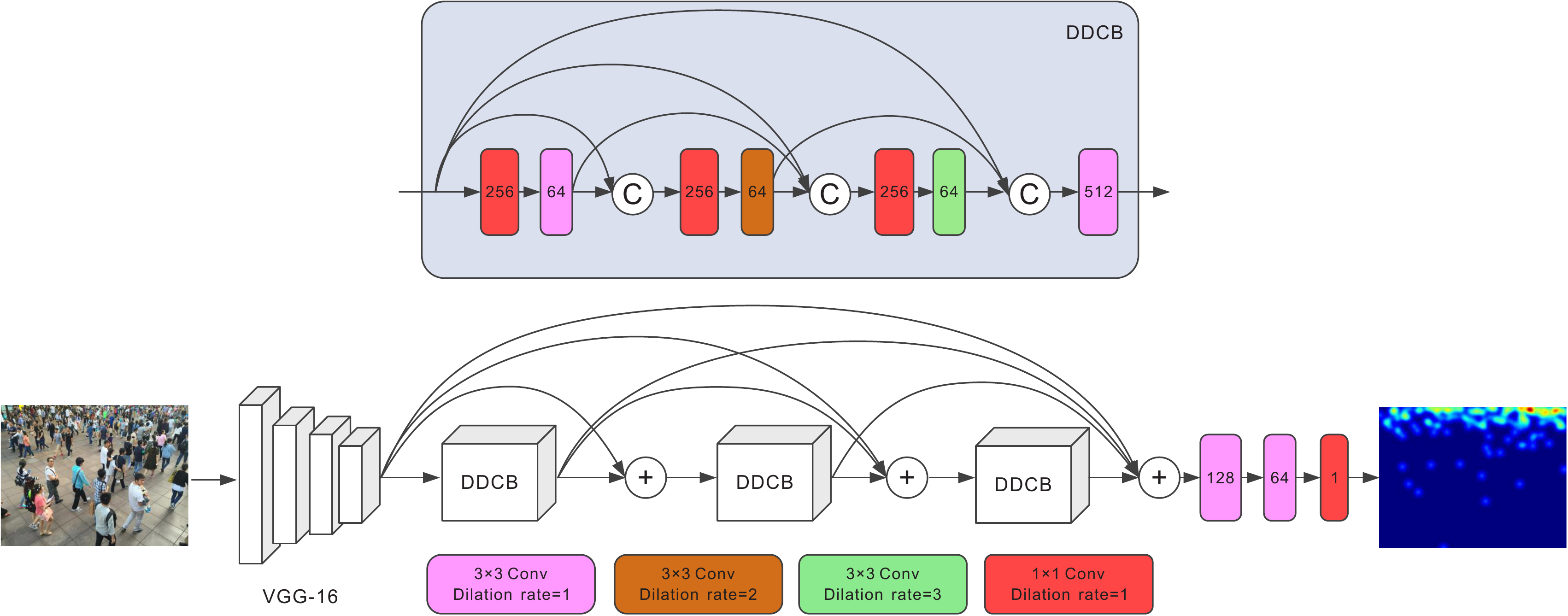}
\end{center}
   \caption{The architecture of the proposed dense scale network (DSNet) for crowd counting. The DSNet consists of backbone network with the front ten layers of VGG-16, three dense dilated convolution blocks (DDCBs) with dense residual connections (DRCs), and three convolutional layers for crowd density map regression. The DDCBs with DRCs are used to enlarge scale diversity and receptive fields of features to handle large scale variations so that density maps can be estimated accurately.}
\label{fig2}
\end{figure*}

\subsection{Traditional methods}
Most of the early traditional works focus on detection-based methods using body or part-based detector to locate people in the crowd image and count their number. However, severe occlusions of highly congested scenes limit the performance of these methods. To overcome the problem, regression-based methods are deployed to learn a mapping from the extracted feature to the number of objects directly. Following similar approaches, Idrees \emph{et al.} \cite{idrees2013multi} proposed a method that extracts features via Fourier analysis and SIFT \cite{lowe1999object} interest points based counting in local patches. Due to the overlooked saliency that causes inaccurate results in local regions, Lempitsky \emph{et al.} \cite{lempitsky2010learning} proposed a method that learns a linear mapping between features and its object density maps in the local region. Futhermore, considering the difficulty of learning an ideal linear mapping, Pham \emph{et al.} \cite{pham2015count} used random forest regression to learn a non-linear mapping instead of the linear one.

\subsection{CNN-based methods}
Due to the success of CNN-based methods in classification and recognition tasks, the CNN-based methods are employed for the purpose of crowd counting and density estimation. Walach \emph{et al.} \cite{walach2016learning} made use of layered boosting and selective sampling methods to reduce the estimation error. Instead of using patch-based training, Shang \emph{et al.} \cite{shang2016end} proposed an estimation method using CNNs which takes the whole image as input and directly outputs the final crowd count. Boominathan \emph{et al.} \cite{boominathan2016crowdnet} presented the first work purely using convolutional network and dual-column architecture to tackle the issue of scale variation for generating density map. Zhang \emph{et al.} \cite{zhang2016single} introduced a multi-column architecture to extract features at different scales. Similarly, Onoro \emph{et al.} \cite{onoro2016towards} proposed a scale-aware, multi-column counting model called Hydra CNN for object density estimation. Recently, inspired by MCNN \cite{zhang2016single}, Sam \emph{et al.} \cite{sam2017switching} proposed a Switching-CNN that adaptively select the most optimal regressor among several independent regressors for a particular patch. Sindagi \emph{et al.} \cite{sindagi2017cnn} explored a new architecture where MCNN \cite{zhang2016single} is enriched with two additional columns capturing global and local context.

Although these multi-column architectures prove the ability to estimate crowd count, several disadvantages also exist in these approaches: they are hard to train caused by the multi-column architecture, and they have large amount of redundant parameters, also the speed is slow as multiple CNNs need to be run. Taking all above drawback into consideration, recent works have focused on multi-scale, single column architectures. Zhang \emph{et al.} \cite{zhang2018crowd} proposed a scale-adaptive CNN that combines adapted feature maps extracted from multiple layers to produce the final density map. Li \emph{et al.} \cite{li2018csrnet} proposed a network for congested scene called CSRNet, which uses dilated kernels to deliver larger reception fields and replace pooling operations. Cao \emph{et al.}  \cite{cao2018scale} presented scale aggregation network that improves the multi-scale representation and generates high-resolution density maps. However, all these single-column works can only capture several kinds of receptive fields, which limits the network to handle large variations in crowd images.

\section{Our Approach}

The fundamental idea of our approach is to deploy an end-to-end single-column CNN with denser scale diversity to cope with the large scale variations and density level differences in both congested and sparse scenes. In this section, we first introduce the architecture of DSNet we proposed as shown in Fig. \ref{fig2}. And then we describe our novel multi-scale density level consistency loss, which enforces the estimated density maps to be consistent with the ground truth according to crowd density at multiple scale levels.

\subsection{DSNet architecture}

Our proposed DSNet contains backbone network as feature extractor, three dense dilated convolution blocks stacked by dense residual connections that enlarges denser scale diversity, and three convolutional layers for crowd density map regression. 

\textbf{Backbone network:} Following CSRNet \cite{zhang2018crowd}, we keep the first ten layers of VGG-16 \cite{simonyan2014very} with only three pooling layers to be our backbone network. According to \cite{simonyan2014very}, using more convolutional layers with small kernels is more efficient than using fewer layers with larger kernels that have been employed in existing multi-column networks. Moreover, it achieves the best tradeoff between accuracy and the resource overhead, which is suitable for accurate and fast crowd counting.

\begin{figure}[t]
	\begin{center}
		\includegraphics[width=0.6\linewidth]{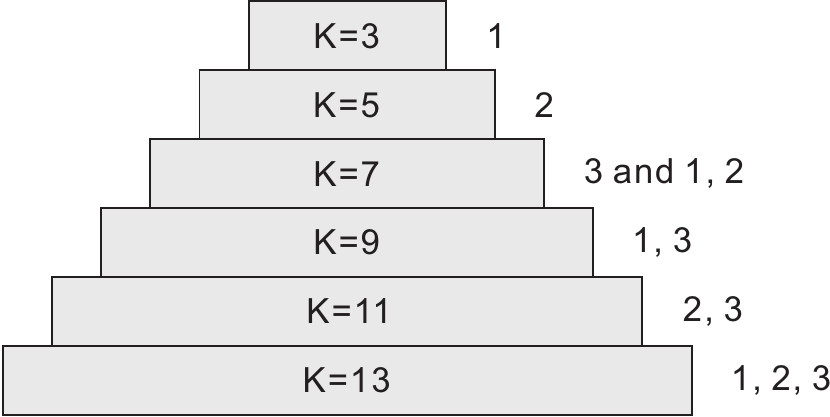}
	\end{center}
	\caption{Illustration of DDCB's scale diversity corresponding to the setting of densely stacked dilation convolutions with dilation rates (1, 2, 3). $K$ represents the receptive field size of the corresponding combination.}
	\label{fig3}
\end{figure}

\textbf{Dense dilated convolution block(DDCB):} As discussed in the first section, the scale often varies continuously across the image and has large range. Considering these challenges, multi-columns CNNs or single-column CNNs with multiple branches have restricted capabilities. This is because they all have limited number of columns or branches, and can only process crowd images with a few different scales. Thus to tackle the challenges of scale variation, we need a network architecture that can capture a large scale range in the manner as dense as possible. DenseASPP \cite{yang2018denseaspp} stacks dilated layers by dense connections to enlarge scale diversity and receptive field for semantic segmentation of high-resolution street images. However, the large dilation rates in their network has led to large gap between the sizes of different receptive field, i.e. 6 pixels. This is not desired for crowd counting. Because the scale variation of crowd scenes caused by camera perspective is nearly continuous, we need more densely sampled scale range. Here, we propose a new dense dilated convolution block that contains three dilated convolutional layers with increasing dilation rate of 1, 2, 3. This setting preserves information from denser scales with small gap of receptive field size, i.e. 2 pixels. Each dilated layer within the block is densely connected with others as in DenseASPP so that each layer can access to all the subsequent layers and pass on information that needs to be preserved. After dense connection, the acquired scale diversity is increased and illustrated in Fig. \ref{fig3}. We further enlarge the scale range by linking multiple blocks with dense residual connections, which will be discussed later.

\begin{figure}[t]
\begin{center}
   \includegraphics[width=0.8\linewidth]{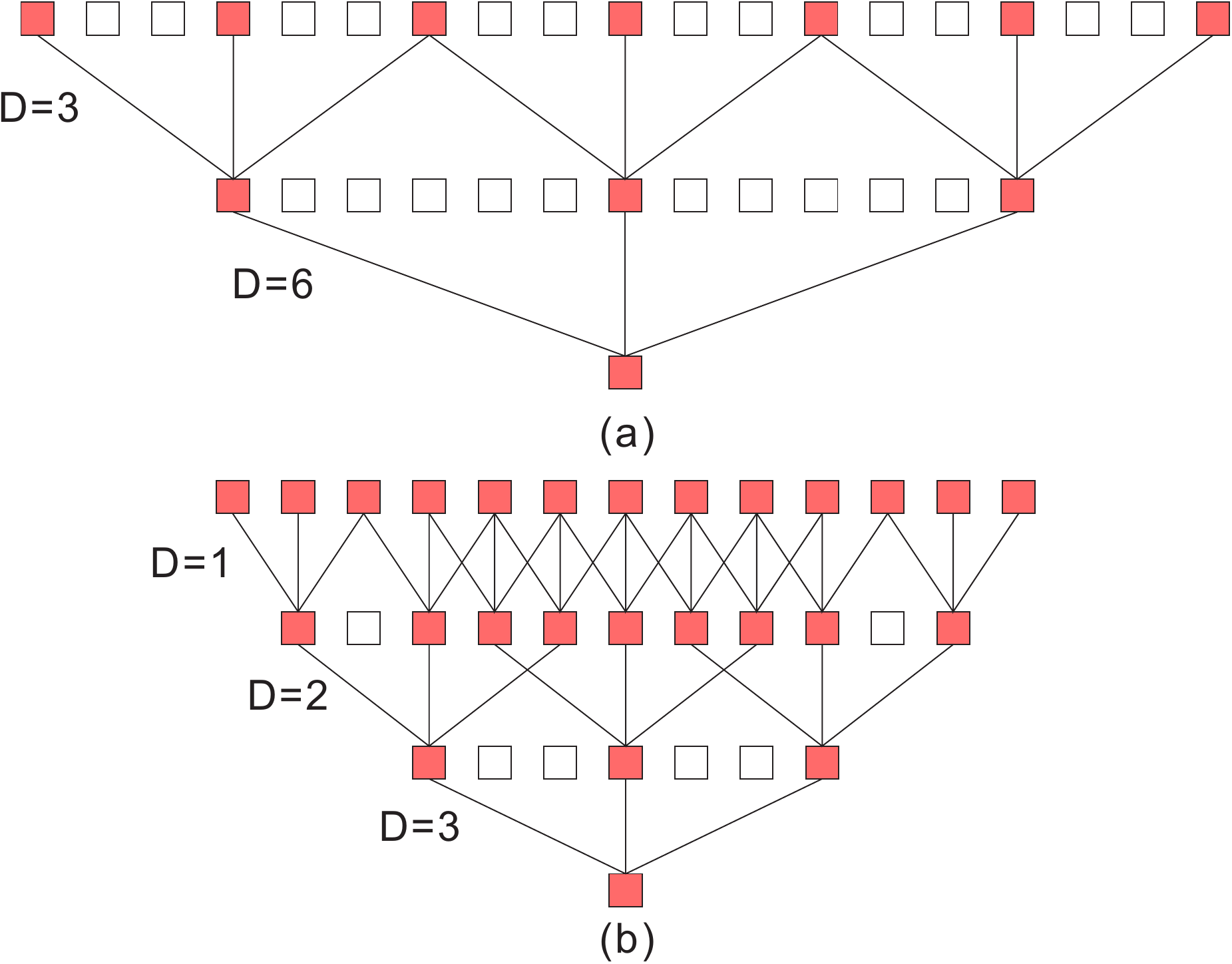}
\end{center}
   \caption{(a) The stacked dilation convolution layers with large dilation rate in DenseASPP, which causes ``gridding artifacts'' that loses a large portion of information. Red color denotes where the information come from. (b) The subsequent convolution layers with dilation rate of (1, 2, 3) in proposed DDCB to cover all pixel information.}
\label{fig4}
\end{figure}

Another advantage of our carefully selected dilation rates is that it can overcome the \textit{gridding artifacts} in DenseASPP \cite{yang2018denseaspp}. As illustrated in Fig. \ref{fig4}, a dilated convolution layer with dilation rate of 6 is put below the layer with dilation rate of 3. The final result of one pixel after these two layers only obtains information from 7 pixels in one-dimensional case. This phenomenon gets worse when the input data is two-dimensional. As a result, the final pixel can only view original information in a gridding fashion and lose a large portion (86.4\%) of information. This is harmful for crowd counting to capture detailed features, as local information of original feature maps is completely missing and the information can be irrelevant across large distances due to large dilation rate. By adopting the new combination of dilation rates, the top layer can cover all pixel information of original feature map and avoid irrelevant information across large distances caused by large dilation rates of intermediate layers. This is critical for crowd counting to estimate accurate density maps. 

Following DenseNet \cite{huang2017densely}, we add a 1 $\times$ 1 convolutional layer before every dilated layer to prevent the block from growing too wide. Moreover, a standard convolution layer with 3 $\times$ 3 filter size is adopted to fuse concated features from front layers and reduce channel number in the end. And ReLU \cite{glorot2011deep} is applied after every convolution layer except the last one.

\textbf{Dense residual connection(DRC):} Although the proposed DDCB provides dense scale diversity, the hierarchical features between different blocks are not fully utilized. Hence, we improve the architecture by dense residual connections to further improve the information flow. Also, they can prevent the network from being wider compared with the conventional dense connections. By doing this, the output of one DDCB has direct access to each layer of the subsequent DDCBs, resulting in a contiguous information pass. Compared with the residual connection without dense style, its scale diversity is further enlarged and the suitable features for specific scenes are preserved adaptively during the flow process of information. The ablation experiments in Section \ref{sec:Ablation} demonstrate that the DRC can improve the performance.

\subsection{Loss function}
Most previous works use Euclidean loss as the loss function for crowd counting, which only takes care of pixel error but ignores the global and local correlations between estimated and ground truth density maps. In this paper, we incorporate multi-scale density level consistency loss that measures global and local context with Euclidean loss.

\textbf{Euclidean loss:} We choose the Euclidean distance to measure the estimation difference at pixel level between the estimated density map and the ground truth, which is similar to other works \cite{li2018csrnet,sam2017switching,boominathan2016crowdnet,zhang2016single}. The loss function is defined as follow:\begin{equation}L_{e}=\frac{1}{N}\sum_{i=1}^{N}\Vert G(X_i;\theta)-D_{i}^{GT}\Vert_{2}^{2}\end{equation}
where $N$ is the number of images in the batch, $G(X_i;\theta)$ is the estimated density map for training image $X_i$ with parameters $\theta$, $D_{i}^{GT}$ is the ground truth density map for $X_i$.

\textbf{Multi-scale density level consistency loss:} Beyond the pixel-wise loss function, we also take the consistency of global and local density level between estimated density maps and ground truth into consideration. The novel proposed training loss is defined as
\begin{equation}L_{c}=\frac{1}{N}\sum_{i=1}^{N}\sum_{j=1}^{S}\frac{1}{k_j^2}\Vert P_{ave}(G(X_i;\theta),k_j)-P_{ave}(D_{i}^{GT},k_j)\Vert_{1}\end{equation}
where $S$ is the number of scale levels for consistency checking, $P_{ave}$ is average pooling operation, $k_j$ is the specified output size of average pooling.

The scale level seperates the density map into different sub-regions and forms pooled representation that illustrates the level of crowd density for different locations. According to the context of density level, the estimated density maps are enforced to be consisted with the ground truth at different scales. Moreover, the number of scale levels and the output size of specific scale can be modified, which provides a tradeoff between the training speed and estimation accuracy. According to our experiments, the performance dosen't improve significantly by adding more additional levels. Therefore we adopt a three level one with output size of 1 $\times$ 1, 2 $\times$ 2 and 4 $\times$ 4 respectively. The first level with output size of 1 $\times$ 1 captures the global context of density level while the other two levels represent the local density level of image patches.

\textbf{Final objective:} By weighting the above two loss functions, the entire network is trained using the following objective function:
\begin{equation} L = L_{e} + \lambda L_{c}\end{equation}
where $\lambda$ is the weight to balance the pixel-wise and density level consistency loss. According to our experiments, its setted values for different datasets are shown in Table \ref{table1}.

\section{Implementation Details}
In this section, we provide specific details of generating ground truth, training process and evaluation process. Due to the good design of network architecture and joint loss function, our network can be trained end-to-end easily.

\subsection{Ground truth generation}

Following the method of generating density maps in \cite{zhang2016single}, the geometry-adaptive kernels are adopted to tackle the datasets with congested scenes including ShanghaiTech Part\_A \cite{zhang2016single}, UCF-QNRF \cite{idrees2018composition} and UCF\_CC\_50 \cite{idrees2013multi}, while the fixed Gaussian kernels are used to generate density maps for the datasets with relatively sparse crowd including ShanghaiTech Part\_B \cite{zhang2016single} and UCSD \cite{chan2008privacy}. By bluring each head annotation using a Gaussian kernel(which is normalized to 1), the ground truth density maps can be generated. The setups of fixed standard deviation for the datasets with sparse crowd follow the configuration in \cite{li2018csrnet}. And the geometry-adaptive kernel used for datasets with congested crowd is defined as follows: 
\begin{equation} D^{GT} = \sum_{i=1}^{N}\delta (x-x_i) \times G_{\sigma_i}(x), with \ \sigma_i = \beta \overline{d_i} \end{equation}
where $x$ is the position of pixel and $N$ is the number of head annotations in the image. For each targeted object $x_i$ in the ground truth $\delta$, use $\overline{d_i}$ to indicate the average distance of $k$ nearest neighbors. Then convolve $\delta(x-x_i)$ with a Gaussian kernel with parameter $\sigma_i$. Following \cite{zhang2016single}, we set $\beta$ = 0.3 and $k$ = 3.

\begin{table}
\begin{center}
\begin{tabular}{|l|c|}
\hline
Dataset & value of weight \\
\hline\hline
ShanghaiTech Part\_A \cite{zhang2016single} & 1000 \\
\hline
ShanghaiTech Part\_B \cite{zhang2016single} & 100 \\
\hline
UCF-QNRF \cite{idrees2018composition} & 1000 \\
\hline
UCF\_CC\_50 \cite{idrees2013multi} & 100\\
\hline
UCSD \cite{chan2008privacy} & 100\\
\hline
\end{tabular}
\end{center}
\caption{The values of weight $\lambda$ for different datasets}
\label{table1}
\end{table}

\begin{table*}
\begin{center}
\begin{tabular}{|l|c|c|c|c|c|c|c|c|c|c|}
\hline
 & \multicolumn{2}{|c|}{ShanghaiTechA} & \multicolumn{2}{|c|}{ShanghaiTechB} & \multicolumn{2}{|c|}{UCF-QNRF} & \multicolumn{2}{|c|}{UCF\_CC\_50}  & \multicolumn{2}{|c|}{UCSD}\\
\hline
Method & MAE & MSE & MAE & MSE & MAE & MSE & MAE & MSE & MAE & MSE\\
\hline\hline
Idress et al. \cite{idrees2013multi} & - & - & - & - & 315.0 & 508.0 & - & - & - & - \\
\hline
Cross-Scene \cite{zhang2015cross} & - & - & - & - & - & - & - & - & 1.60 & 3.31 \\
\hline
MCNN \cite{zhang2016single} & 110.2 & 173.2 & 26.4 & 41.3 & 277.0 & 426.0 & 377.6 & 509.1 & 1.07 & 1.35 \\
\hline
C-MTL \cite{sindagi2017cnn} & 101.3 & 152.4 & 20.0 & 31.1 & 252.0 & 514.0 & 322.8 & 341.4 & - & - \\
\hline
SwitchCNN \cite{sam2017switching} & 90.4 & 135.0 & 21.6 & 33.4 & 228.0 & 445.0 & 318.1 & 439.2 & 1.62 & 2.10 \\
\hline
SaCNN \cite{zhang2018crowd} & 86.8 & 139.2 & 16.2 & 25.8 & - & - & 314.9 & 424.8 & - & - \\
\hline
CP-CNN \cite{sindagi2017generating} & 73.6 & 106.4 & 20.1 & 30.1 & - & - & 295.8 & 320.9 & - & - \\
\hline
ACSCP \cite{shen2018crowd} & 75.7 & 102.7 & 17.2 & 27.4 & - & - & 291.0 & 404.6 & 1.04 & 1.35 \\
\hline
Deep-NCL \cite{shi2018crowd} & 73.5 & 112.3 & 18.7 & 26.0 & - & - & 288.4 & 404.7 & - & - \\
\hline
IG-CNN \cite{babu2018divide} & 72.5 & 118.2 & 13.6 & 21.1 & - & - & 291.4 & 349.4 & - & - \\
\hline
ic-CNN \cite{ranjan2018iterative} & 68.5 & 116.2 & 10.7 & 16.0 & - & - & 260.9 & 365.5 & - & - \\
\hline
CSRNet \cite{li2018csrnet} & 68.2 & 115.0 & 10.6 & 16.0 & - & - & 266.1 & 397.5 & 1.16 & 1.47 \\
\hline
CL-CNN \cite{idrees2018composition} & - & - & - & - & 132.0 & 191.0 & - & - & - & - \\
\hline
SANet \cite{cao2018scale} & 67.0 & 104.5 & 8.4 & 13.6 & - & - & 258.4 & 334.9 & 1.02 & 1.29 \\
\hline
DSNet & \textbf{61.7} & \textbf{102.6} & \textbf{6.7} & \textbf{10.5} & \textbf{91.4} & \textbf{160.4} & \textbf{183.3} & \textbf{240.6} & \textbf{0.82} & \textbf{1.06} \\
\hline
\end{tabular}
\end{center}
\caption{Comparison with state-of-the-art methods on ShanghaiTech \cite{zhang2016single}, UCF-QNRF \cite{idrees2018composition}, UCF\_CC\_50 \cite{idrees2013multi} and UCSD \cite{chan2008privacy} datasets. Our approach achieves the best performance by a large margin compard with the state-of-the-art methods. All results of previous methods are cited from original papers.}
\label{table2}
\end{table*}

\subsection{Training details}
\label{sec42}
Similar to other recent crowd counting works \cite{boominathan2016crowdnet,sam2017switching,sindagi2017generating,li2018csrnet}, the backbone network with ten layers is fine-tuned from a well-trained VGG-16 \cite{simonyan2014very}. All new layers are initialized from a Gaussian distribution with zero mean and 0.01 standard deviation. Adam \cite{kingma2014adam} optimizer is applied with fixed learning rate at 5e-6 and weight decay of 5e-4 because it shows faster convergence than stochastic gradient descent with momentum. And the network is trained with batch size of 1. The implementation of our method is based on the Pytorch \cite{paszke2017pytorch} framework.

Furthermore, during training, we crop image patch of 1/4 size of the original image at four quarters of the image without overlapping and other patches are randomly cropped from the image. After that, it is horizontal flipped randomly with probability 0.5. Taking illumination variation into consideration, we adopt gamma transform using parameter [0.5, 1.5] with probability 0.3 for ShanghaiTech \cite{zhang2016single} and UCF-QNRF \cite{idrees2018composition} dataset. Also, we also randomly change the color images to gray with probability 0.1 for the ShanghaiTech Part\_A \cite{zhang2016single} and UCF-QNRF \cite{idrees2018composition} datasets that contain gray images. 

Moreover, the image resolution of UCF-QNRF \cite{idrees2018composition} dataset is larger than all other datasets, causing GPU out of memory in the training process. We resize large images to a maximum size of 720p before data augmentation. On the contrary, the image resolution of UCSD \cite{chan2008privacy} dataset is 238 $\times$ 158, which is too small to generate high-quality density maps. Hence, we enlarge each image to 952 $\times$ 632 resolution. 

\subsection{Evaluation details}

At test time, we do not crop image patches from the images and instead we feed the whole image into the network to generate the estimated density maps. And the mean absolute error (MAE) and mean square error (MSE) are adopted to evaluate the performance of our network. Moreover, the MAE reflects the model's accuracy while the MSE reflects the model's robustness. And lower value means better performance. These two metrics are defined as follows:
\begin{equation} MAE = \frac{1}{N}\sum_{i=1}^N |C_i-C_i^{GT}|\end{equation}
\begin{equation} MSE = \sqrt{\frac{1}{N}\sum_{i=1}^N |C_i-C_i^{GT}|^2}\end{equation}
where $N$ is the number of images in test set, $C_i$ represents the estimated count while $C_i^{GT}$ is the ground truth count.

\section{Experiments}

In this section, we introduce four publicly available datasets used to evaluate our approach firstly. Then we evaluate and compare our proposed approach to the previous state-of-the-art methods on these four datasets. Futhermore, ablation experiments on the ShanghaiTech Part\_B dataset is conducted to demonstrate the effectiveness of every module in our network architecture and proposed density level consistency loss.

\begin{figure*}
\begin{center}
\begin{minipage}{4cm}
\includegraphics[width=4cm, height=3cm]{img1}
\end{minipage}
\begin{minipage}{4cm}
\includegraphics[width=4cm, height=3cm]{img2}
\end{minipage}
\begin{minipage}{4cm}
\includegraphics[width=4cm, height=3cm]{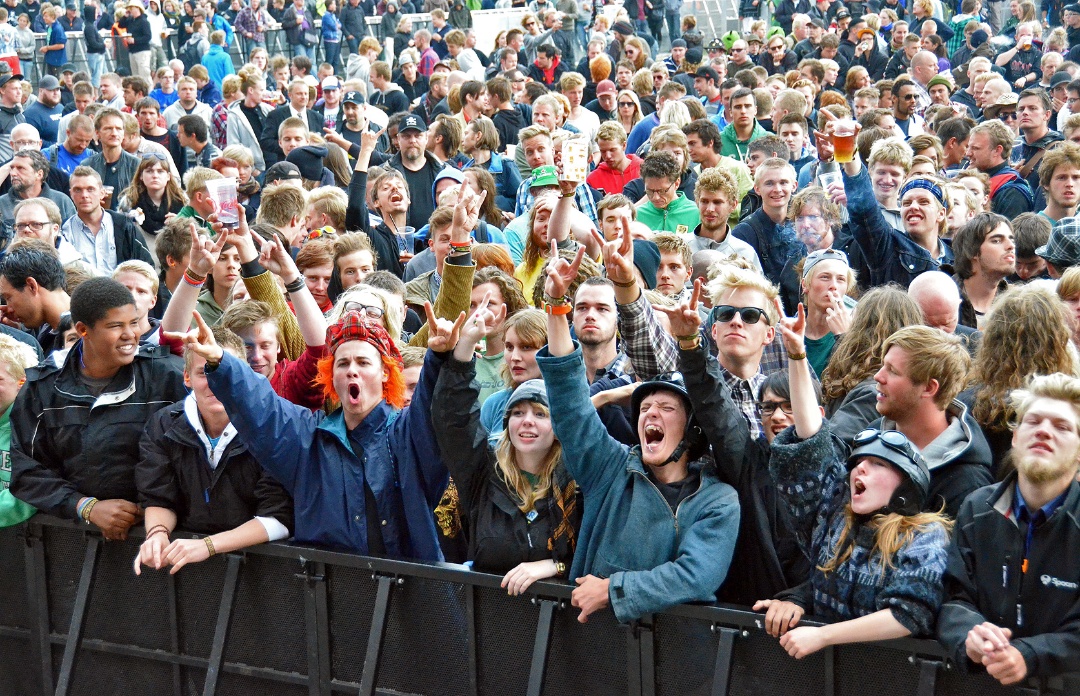}
\end{minipage}
\begin{minipage}{4cm}
\includegraphics[width=4cm, height=3cm]{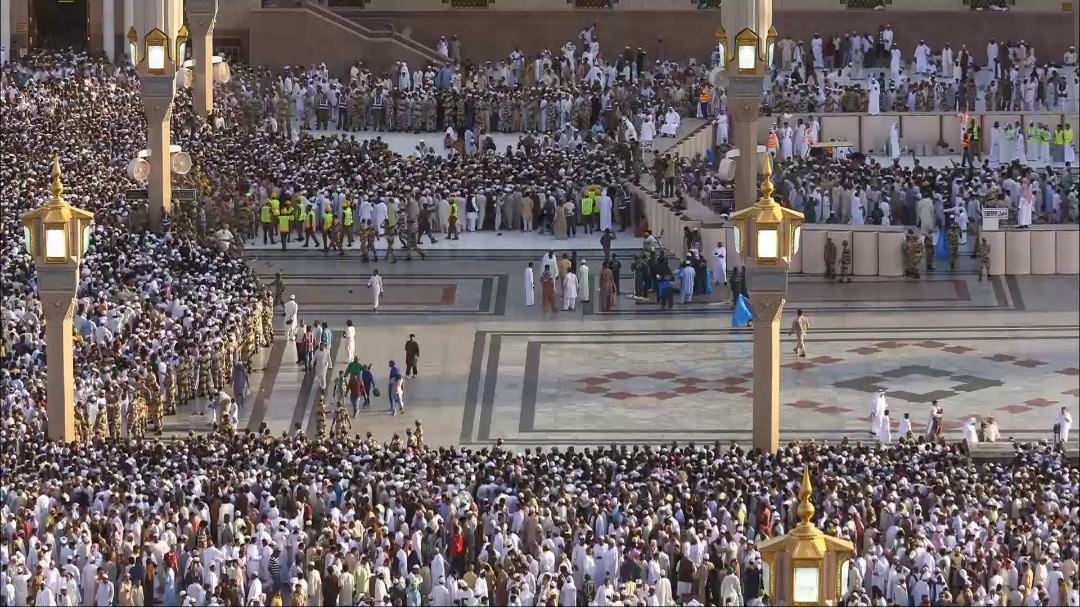}
\end{minipage}

\begin{minipage}{4cm}
\includegraphics[width=4cm, height=3cm]{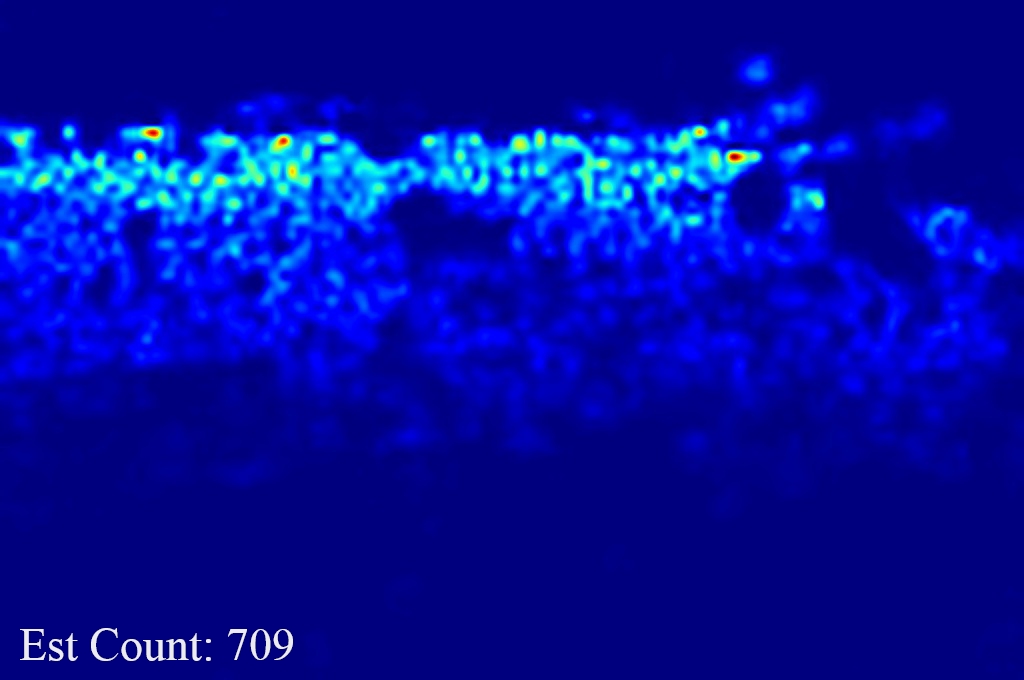}
\end{minipage}
\begin{minipage}{4cm}
\includegraphics[width=4cm, height=3cm]{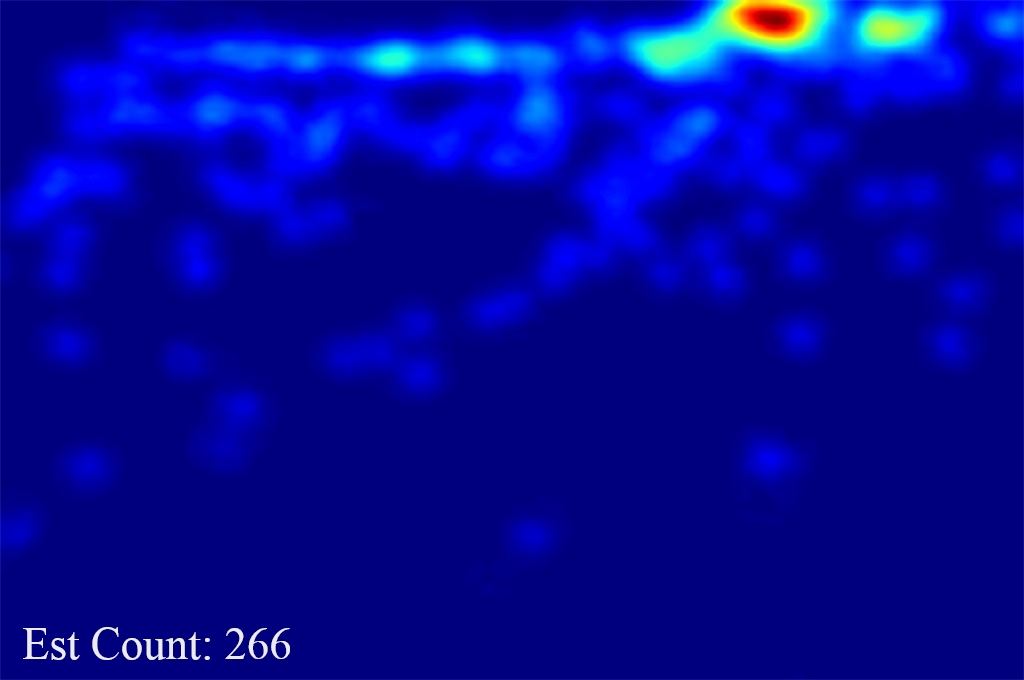}
\end{minipage}
\begin{minipage}{4cm}
\includegraphics[width=4cm, height=3cm]{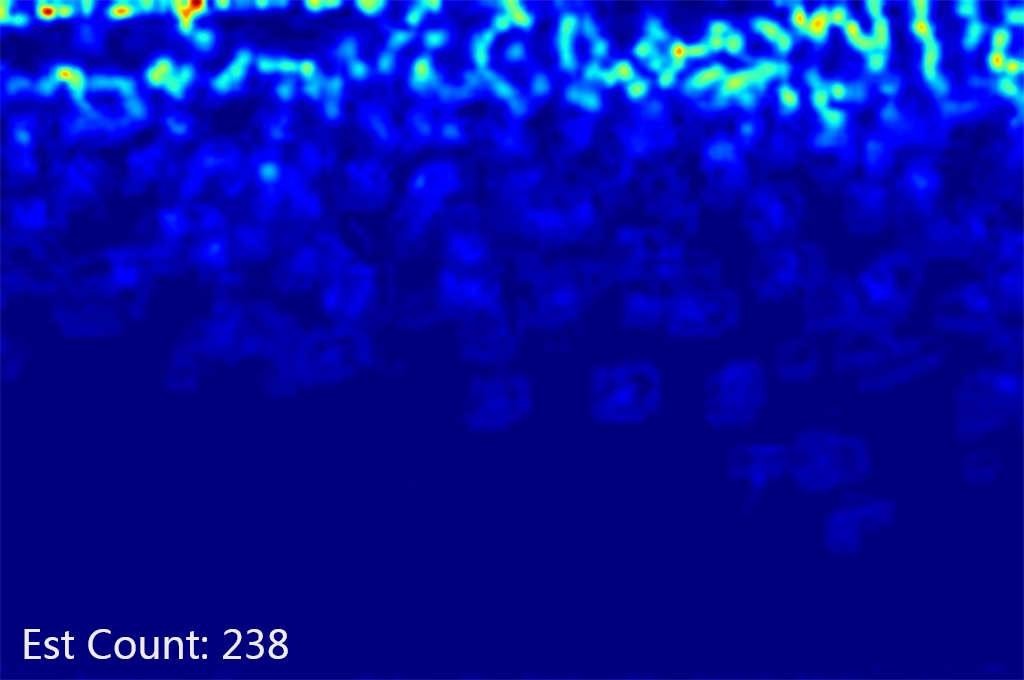}
\end{minipage}
\begin{minipage}{4cm}
\includegraphics[width=4cm, height=3cm]{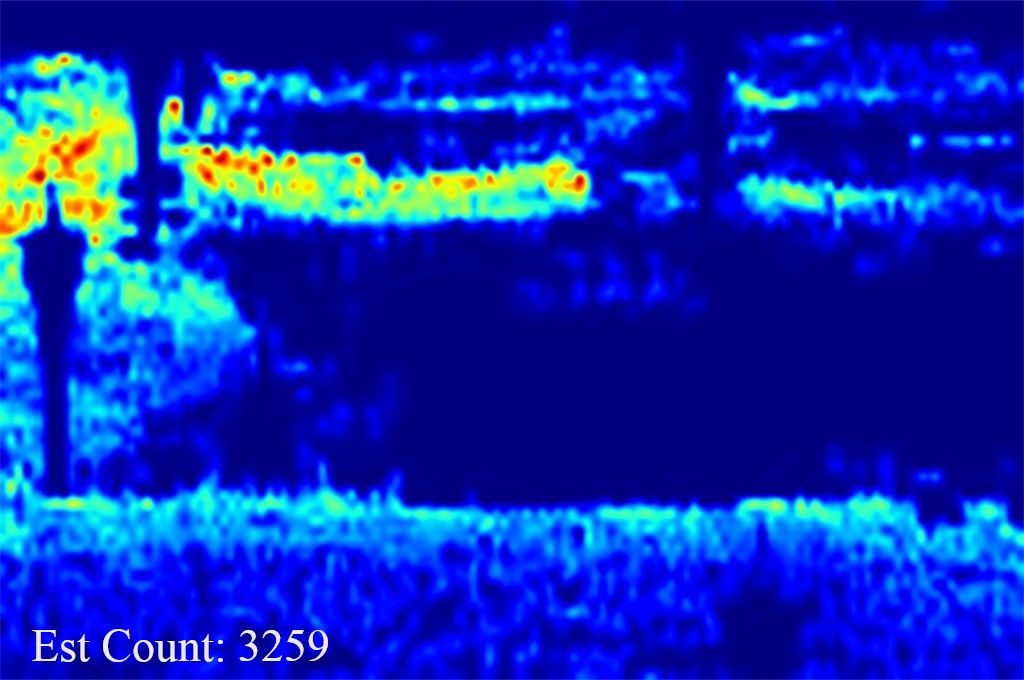}
\end{minipage}

\begin{minipage}{4cm}
\includegraphics[width=4cm, height=3cm]{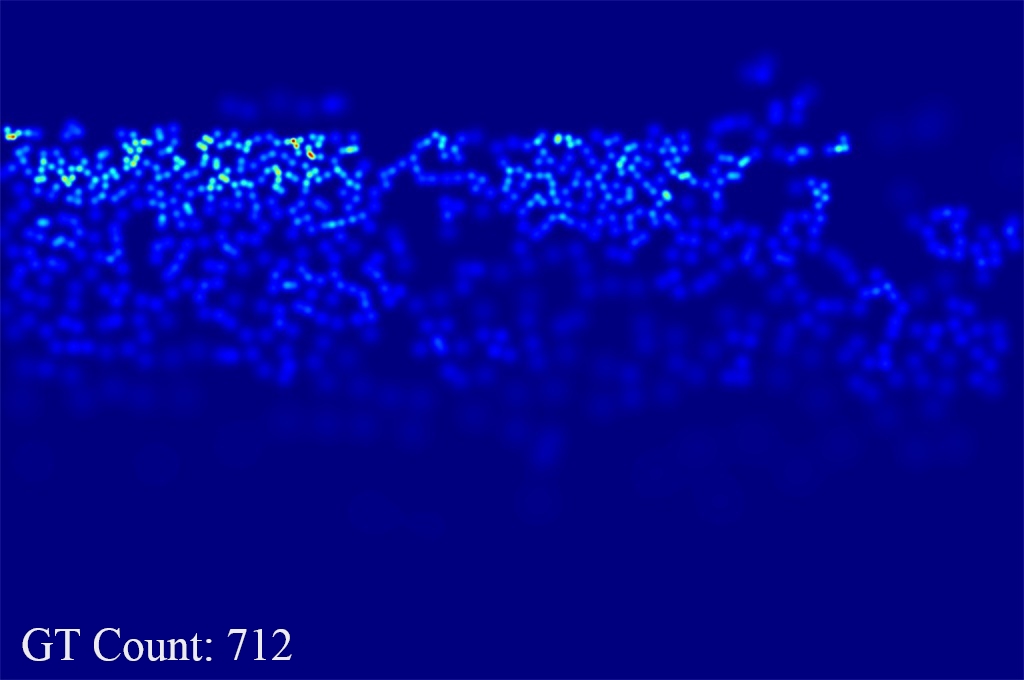}
\end{minipage}
\begin{minipage}{4cm}
\includegraphics[width=4cm, height=3cm]{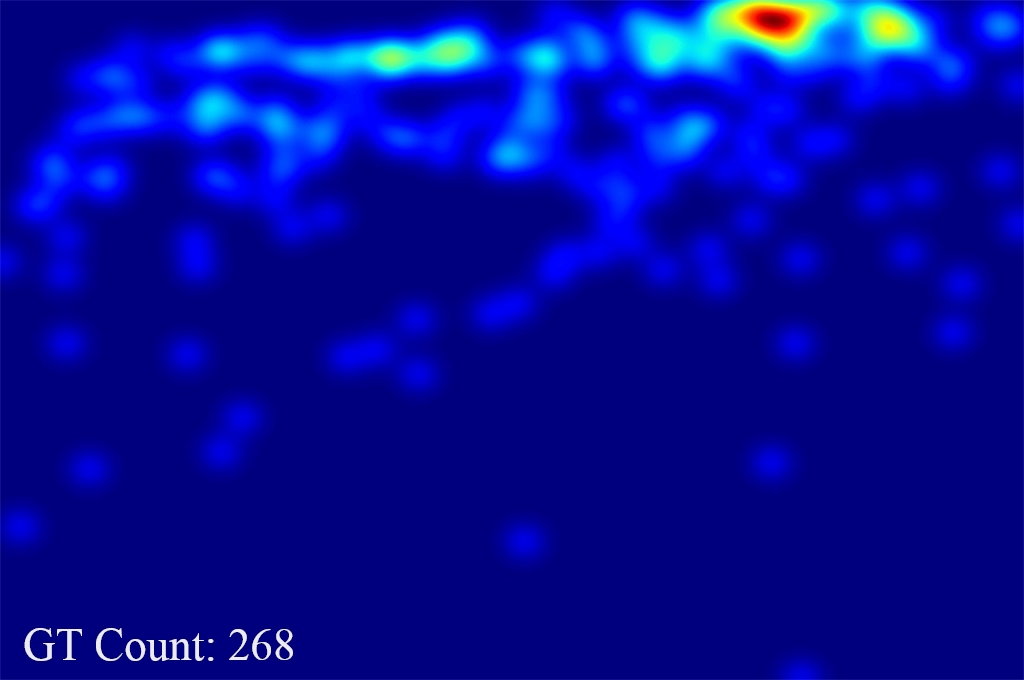}
\end{minipage}
\begin{minipage}{4cm}
\includegraphics[width=4cm, height=3cm]{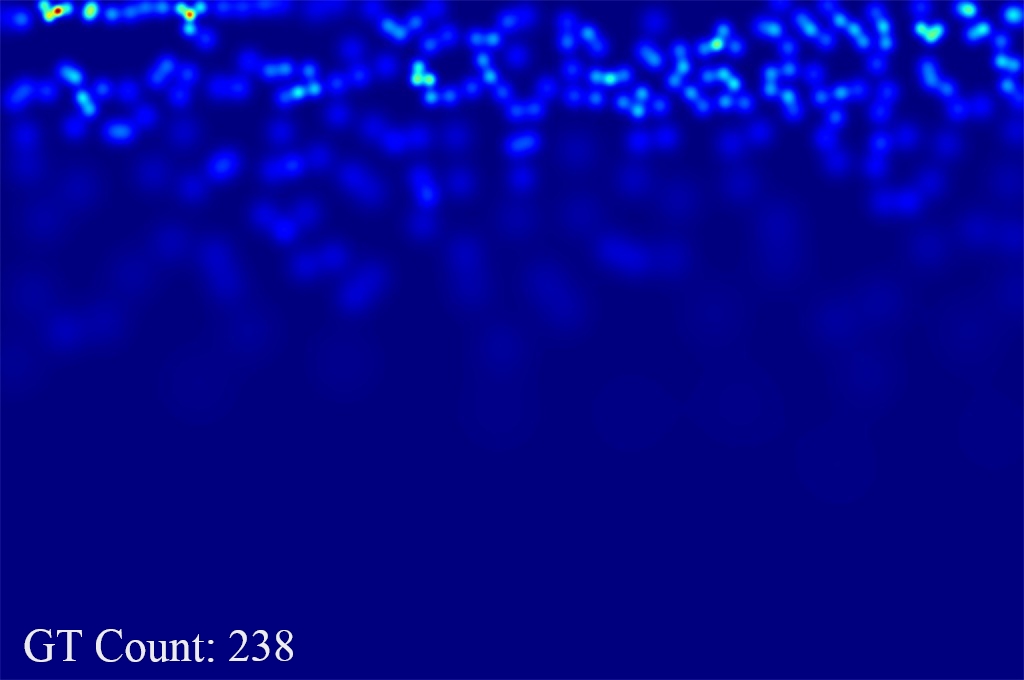}
\end{minipage}
\begin{minipage}{4cm}
\includegraphics[width=4cm, height=3cm]{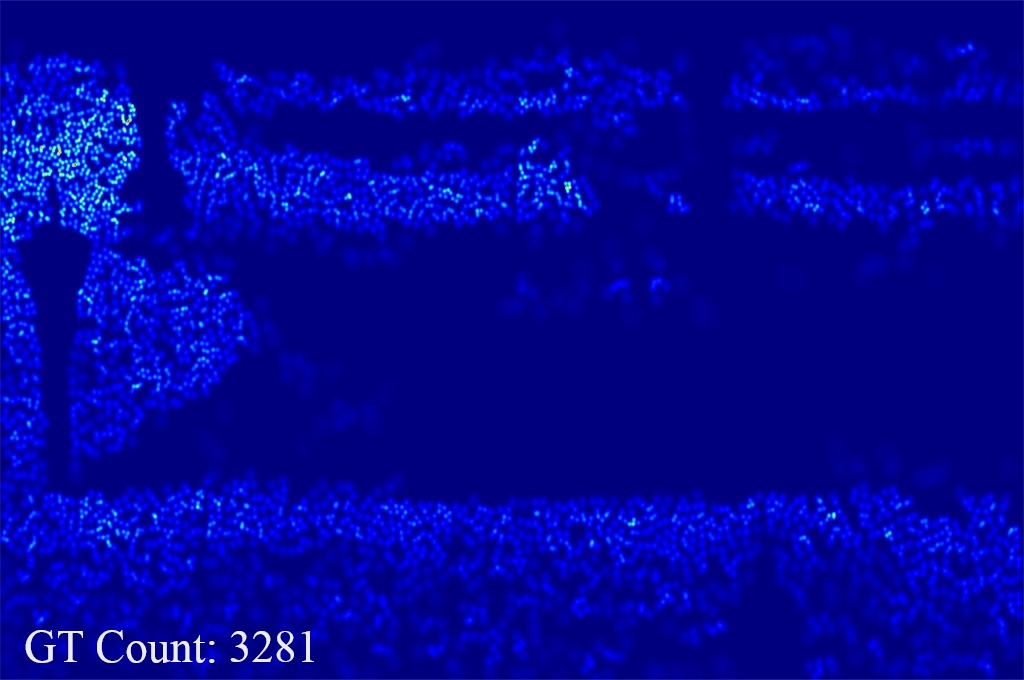}
\end{minipage}
\end{center}
\caption{An illustration of estimated density maps and crowd counts generated by proposed DSNet. The first row shows four samples drawn from ShanghaiTech Part\_A, ShanghaiTech Part\_B and UCF-QNRF datasets. The second row shows the density maps estimated by DSNet. And the last row shows the corresponding ground truth maps. DSNet generates density maps close to the ground truth and accurate crowd counts.}
\label{fig5}
\end{figure*}

\subsection{Datasets}
We evaluate our DSNet on four pulicly available crowd counting datasets: ShanghaiTech \cite{zhang2016single}, UCF-QNRF \cite{idrees2018composition}, UCF\_CC\_50 \cite{idrees2013multi} and UCSD \cite{chan2008privacy}.

\textbf{ShanghaiTech:} It contains 1198 annotated images with a total amount of 330,165 persons \cite{zhang2016single}, which is divided into part A and part B. Pat A contains 482 images with highly congested scenes that counts varies from 33 to 3139 randomly downloaded from the Internet.Its training set has 300 images and the testing set has 182 images not in the training set. And part B includes 716 images with relatively sparse crowd scenes taken from fixed cameras of streets that counts varies from 12 to 578. Equally, its training set contains 400 images and the testing set has 316 images. 

\textbf{UCF-QNRF:} It is a latest released and the largest crowd dataset \cite{idrees2018composition}. It consists of 1535 dense crowd images from Flickr, Web Search and Hajj footage. The dataset has a wider variety of scenes containing the most diverse set of viewpoints, lighting variations and densities that counts varies from 49 to 12865, which makes it more difficult and realistic. Moreover, the image resolution is also very large leading to the drastic variation of the size of heads.

\textbf{UCF\_CC\_50:} It includes 50 black and white, low resolution images with extremely dense crowd scenes \cite{idrees2013multi}. The number of annotated persons per image ranges from 94 to 4543 with an average number of 1280, which makes it challenging for a deep-learning approach. 

\textbf{UCSD:} It consists of 2000 frames with size of 238 $\times$ 158 captured by surveillance cameras \cite{chan2008privacy}. This dataset has relatively low density varying from 11 to 46 persons per image with an average of around 25 people. Among all frames, frames 601 through 1400 are taken as training set and the rest of them forms testing set following \cite{chan2008privacy}. And all frames and density maps are masked with provided ROI.

\subsection{Comparisons with State-of-the-Art}
\label{sec52}
To demonstrate the effectiveness of our proposed approach, we conduct comparative experiments on four public challenging crowd counting datasets. The experimental results are shown in Table \ref{table2} for ShanghaiTech, UCF-QNRF, UCF\_CC\_50 and UCSD respectively. It is obvious that our proposed method achieves state-of-the-art performance compared with the previous methods on all datasets and all evaluation metrics, which indicates that our method can perform very well not only on the congested crowd scenes but also the sparse crowd scenes.

For ShanghaiTech datast, our model achieves the lowset MAE and MSE in Part\_A compared to other methods and we get 7.9\% MAE improvement for Part\_A compared with the state-of-the-art method SANet. Also, DSNet delivers 20.2\% lower MAE and 22.8\% lower MSE in Part\_B compared with SANet. On UCF-QNRF dataset, which is the largest dataset and the one with the largest variation of head size, our model also attains the best performance with 30.8\% and 16.0\% improvement for MAE and MSE respectively compared with the second best approach. It illustrates the ability of DSNet to handle large scale variations. On UCF\_CC\_50 dataset, DSNet achieves 29.1\% MAE improvement compared with SANet and 25.0\% MSE improvement compared with CP-CNN. Finally, it has 19.6\% MAE and 17.8\% MSE improvement for the UCSD dataset compared with SANet, which demonstrates that our approach not only achieves superior performance on congested crowd scenes but also on sparse crowd dataset. 

Finally, several examples of our approach are presented in Fig.\ref{fig5}. It is obvious that our method performs well on counting the number of people in images. It also validates that the large variated heads with different sizes can be captured so that DSNet becomes more robust and accurate for not only congested crowd scenes but also sparse scenes.

\subsection{Ablation experiments}\label{sec:Ablation}

In this section we conduct ablation experiments of our network components as well as the configuration of loss function and analyze the outputs of experiments. All these experiments are conducted on the ShanghaiTech Part\_B dataset because it is captured by surveillance cameras of realistic streets, which can demonstrate the ability to cope with real scenes. 

\textbf{Network architecture:} Our proposed approach contains backbone network, dense dilated convolution block, dense residual connection and multi-scale density level consistency loss. To demonstrate their effectiveness, we conduct experiments by adding these components incrementally. And the experimental results are shown in Table \ref{table3}.

We use the backend network and the last three convolution layers to be the baseline model. It achieves an MAE of 15.21, which is the lowest across all entries in the table but can still be comparable to a majority of existing methods.

\begin{table}
\begin{center}
\begin{tabular}{|l|c|c|}
\hline
Method & MAE & MSE \\
\hline\hline
VGG-16 & 15.21 & 22.96\\
\hline
VGG-16+DDCB(1) & 10.71 & 15.89\\
\hline
VGG-16+DDCB(2) & 8.67 & 13.76\\
\hline
VGG-16+DDCB(3) & 7.33 & 12.07\\
\hline
VGG-16+DDCB(3)+DRC & 7.06 & 12.01\\
\hline
VGG-16+DDCB(3)+DRC+$L_c$ & \textbf{6.74} & \textbf{10.48}\\
\hline
\end{tabular}
\end{center}
\caption{Estimation errors for different components of our proposed network on ShanghaiTech Part\_B \cite{zhang2016single}. The number in bracket is the number of dense dilated convolution block.}
\label{table3}
\end{table}

\begin{table}
\begin{center}
\begin{tabular}{|l|c|c|}
\hline
Method & MAE & MSE \\
\hline\hline
w/o Residual Connection & 6.86 & 10.85\\
\hline
Residual Connection & 6.81 & 10.54\\
\hline
Dense Residual Connection & \textbf{6.74} & \textbf{10.48}\\
\hline
\end{tabular}
\end{center}
\caption{Estimation errors for different configurations of residual connection on ShanghaiTech Part\_B \cite{zhang2016single}.}
\label{table4}
\end{table}

By only adding the proposed DDCB incrementally to enrich the baseline model, the MAE decreases to 7.33, which improves by a big margin and achieves the best performance compared with pervious methods. This illustrates that the features with dense scales and large recptive fields caused by incremental dense dilated convolution block are essential and beneficial to count crowd accurately and robustly. 

Furthermore, the dense residual connections are added between three dense dilated convolution block. They also bring an improvement and the MAE further decreases to 7.06, which indicates that the dense residual connections enlarge the scale diversity furtherly by reusing features from different DDCB.

Finally, add our density level consistency loss to train the whole network. It further decreases the mean absoluate error to 6.74, which is the best performance of our approach and achieves the state-of-the-art on the dataset. It demonstrates that the propoesd loss can enforce the density level of estimated density map to be consistent with ground truth at global and local level.

Moreover, we also compare the influence of dense residual connections with ordinary residual connections. The experimental results are reported in Table \ref{table4}. By utilizing residual connections, the estimated error decreases to 6.81 due to the reused features from the previous one block while ignoring the features from other blocks with different scales. To address the issue, dense residual connections are adopted to further decrease the MAE to 6.74, which indicates that scale diversity are further enlarged and features are more effective.

\begin{table}
\begin{center}
\begin{tabular}{|l|c|c|}
\hline
Method & MAE & MSE \\
\hline\hline
w/o $L_c$ & 7.06 & 12.01\\
\hline
Level 1 & 6.95 & 11.38\\
\hline
Level 1,2 & 6.88 & 11.35\\
\hline
Level 1,2,4 & \textbf{6.74} & \textbf{10.48}\\
\hline
\end{tabular}
\end{center}
\caption{Estimation errors for different levels of our proposed consistency loss on ShanghaiTech Part\_B \cite{zhang2016single}. The number is the output size of average pooling operation.}
\label{table5}
\end{table}

\textbf{Loss function:} Our proposed novel loss adopts a configuration with three scale levels (i.e. 1 $\times$ 1, 2 $\times$ 2, 4 $\times$ 4 output size of average pooling operation). We conduct experiments with these three levels to present that every scale level can regularize the consistence between the estimated density map and ground truth. Experimental results are presented in Table \ref{table5}.

Before we add the consistency loss function, the proposed network achieves an MAE of 7.06. By adopting single level with the output size of 1 $\times$ 1, which is the global context that represents the density level of the whole input image, the mean absolute error decreases to 6.95. Furthermore, the performance is continued to be improved due to the influence of the local levels with the output size of 2 $\times$ 2 and 4 $\times$ 4 that decreases the MAE to 6.88 and 6.74 individually. All these incremental experiments indicates that both global and local regularization of density levels can help the estimated density maps to be consisted with the ground truth maps at different scale levels so that high-quality density map can be generated.

\section{Conclusion}

In this paper, we propose a novel end-to-end single-column model called DSNet to estimate crowd count accurately, based on the dense dilated convolution blocks with dense residual connections. These two components enlarge scale diversity and receptive field of features that can handle the issue of large scale variations to perform well on counting the number of people in images. We further introduce a novel loss to enforce the density level of estimated density maps to be consisted with corresponding ground truth maps at different scale levels. Our proposed approach achieves state-of-the-art results on four public challenging crowd counting datasets, on all evaluation metrics.

{\small
\bibliographystyle{ieee}
\bibliography{egbib}
}

\end{document}